\documentclass[sigconf]{acmart}

\AtBeginDocument{%
  }

\copyrightyear{2024}
\acmYear{2024}
\setcopyright{acmlicensed}\acmConference[MMSports '24]{Proceedings of the 7th ACM International Workshop on Multimedia Content Analysis in Sports}{October 28-November 1 2024}{Melbourne, VIC, Australia}
\acmBooktitle{Proceedings of the 7th ACM International Workshop on Multimedia Content Analysis in Sports (MMSports '24), October 28-November 1 2024, Melbourne, VIC, Australia}
\acmDOI{10.1145/3689061.3689077}
\acmISBN{979-8-4007-1198-5/24/10}

\usepackage{subcaption}
\usepackage{booktabs}

\begin{document}

\title[3D Pose-Based Temporal Action Segmentation for Figure Skating]{3D Pose-Based Temporal Action Segmentation for Figure Skating: A Fine-Grained and Jump Procedure-Aware Annotation Approach}


\author{Ryota Tanaka}
\affiliation{%
  \institution{Nagoya University}
  \city{Nagoya}
  \country{Japan}}
\email{tanaka.ryota@g.sp.m.is.nagoya-u.ac.jp}

\author{Tomohiro Suzuki}
\affiliation{%
  \institution{Nagoya University}
  \city{Nagoya}
  \country{Japan}}
\email{suzuki.tomohiro@g.sp.m.is.nagoya-u.ac.jp}

\author{Keisuke Fujii}
\affiliation{%
  \institution{Nagoya University}
  \city{Nagoya}
  \country{Japan}}
\email{fujii@i.nagoya-u.ac.jp}


\renewcommand{\shortauthors}{Ryota Tanaka, Tomohiro Suzuki, \& Keisuke Fujii}

\begin{abstract}
Understanding human actions from videos is essential in many domains, including sports. In figure skating, technical judgments are performed by watching skaters' 3D movements, and its part of the judging procedure can be regarded as a Temporal Action Segmentation (TAS) task. TAS tasks in figure skating that automatically assign temporal semantics to video are actively researched. However, there is a lack of datasets and effective methods for TAS tasks requiring 3D pose data. In this study, we first created the FS-Jump3D dataset of complex and dynamic figure skating jumps using optical markerless motion capture. We also propose a new fine-grained figure skating jump TAS dataset annotation method with which TAS models can learn jump procedures. In the experimental results, we validated the usefulness of 3D pose features as input and the fine-grained dataset for the TAS model in figure skating. FS-Jump3D Dataset is available at \url{https://github.com/ryota-skating/FS-Jump3D}.
\end{abstract}





\maketitle

\section{Introduction}

\begin{figure*}[ht]
    \centering
    \includegraphics[width=0.9\linewidth]{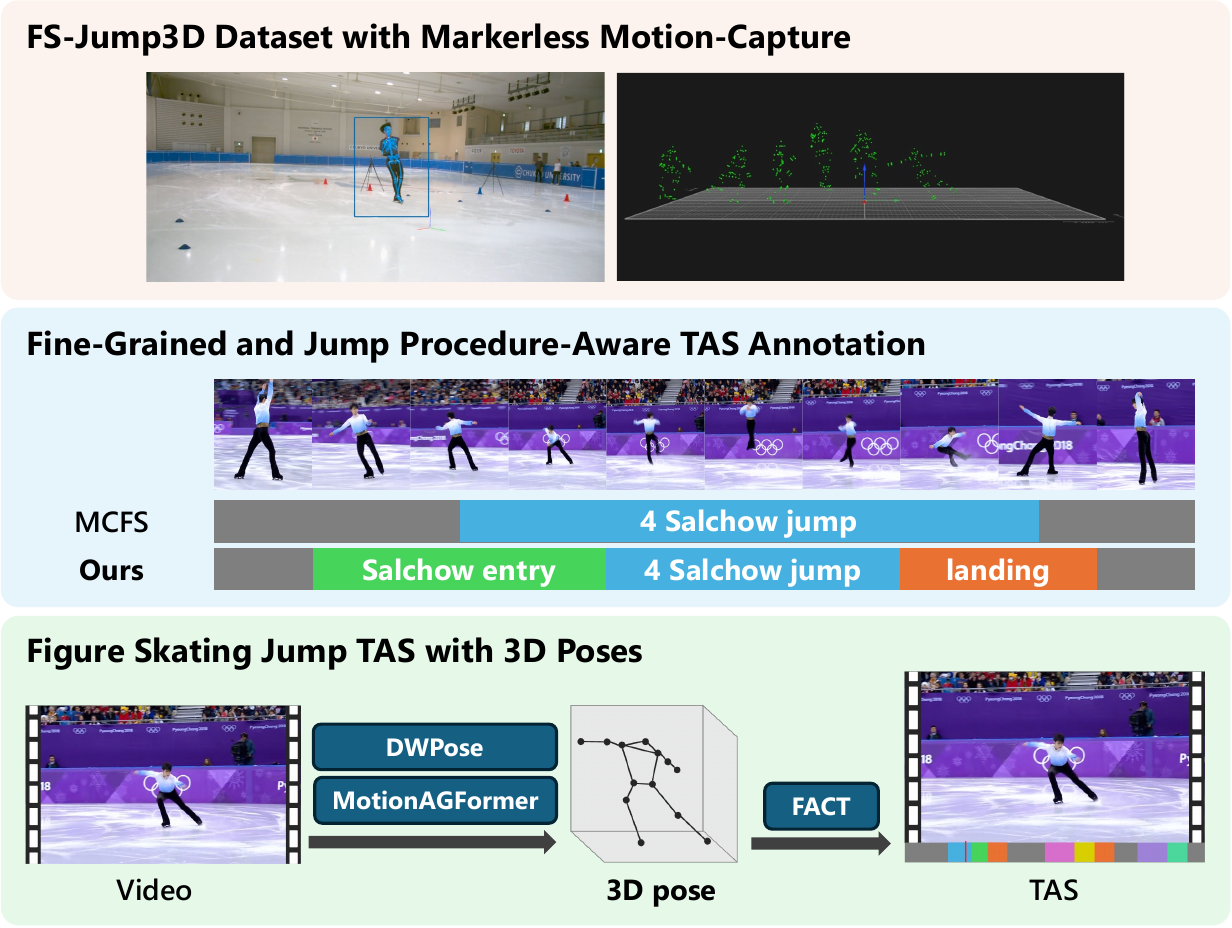}
    \caption{The overview of our proposed method.  First, we created the FS-Jump3D dataset of complex and dynamic figure skating jumps using optical markerless motion capture. Second, we propose a new fine-grained figure skating jump TAS dataset annotation (the difference from \cite{MCFS} is shown in the figure). Lastly, we estimate the 3D poses from the broadcast video dataset with DWPose \cite{DWPose} and MotionAGFormer \cite{motionagformer2024}, and perform TAS with FACT \cite{FACT} on the dataset annotated with detailed figure skating jump procedures.}
    \label{fig:overview}
\end{figure*}

Understanding human actions from videos is crucial in various fields, including sports \cite{suzuki2024automatic, tanaka2023automatic_mmsports}, autonomous driving \cite{action_recog_pedestrian}, security \cite{action_recog_secure}, and elderly care \cite{action_recog_nursing}. In sports, video data is often used for replay judgment and feedback during practice. In figure skating, technical judgments are becoming more challenging as athletes' skills improve, making video data indispensable. Currently, technical specialists and replay operators manually record the type and timing of jumps for video replay after the performance, but the process requires specialized knowledge and is costly. This study aims to address this issue by developing a solution for the Temporal Action Segmentation (TAS) task, which automatically annotates the type and timing of jumps at the frame level using time-series features from performance videos.

TAS involves dividing untrimmed video data into action segments at the frame level in the temporal direction \cite{TAS_analysis}. The TAS datasets are mainly dealing with cooking procedures in the kitchen \cite{Breakfast, GTEA, 50Salads, Epic-Kitchen} or assembly procedures of furniture and toys \cite{Ego-ProceL, Assembly101, Ikea_ASM, Meccano}, with actions such as ``open'' and ``close'' annotated for each frame, for example. The TAS model is designed to understand the time-series ``procedures'' in these kitchen datasets and furniture assembly datasets. TAS models have been extensively studied, and approaches that separate feature extraction and TAS tasks into two steps have shown high performance \cite{FACT, ASformer, MSTCN, MSTCN++}. Selecting an appropriate feature extraction method is essential to fully utilizing the model's capabilities. In general TAS tasks, image features such as those from I3D \cite{I3D} have been used as input features because information on tools and objects in video frames can help understand human actions.

On the one hand, previous research on figure skating \cite{MCFS, VPD} has demonstrated that 2D poses and embedded pose representations aid in understanding figure skating jump movements, independent of tools and objects in the video frame. However, figure skating movements are inherently three-dimensional, and the varying camera angles in broadcast videos cause additional challenges. Consequently, depth-aware 3D pose features, derived from monocular 3D pose estimation in broadcast videos, are expected to be highly beneficial. Despite this, no studies have validated their effectiveness. Moreover, earlier studies \cite{MCFS, SkatingVerse} on figure skating TAS tasks have not accounted for jump preparation and landing movements in their action label annotations. Since TAS models are designed to comprehend procedural actions, it will be essential to incorporate jump procedures in figure skating TAS tasks. Therefore, this study proposes a fine-grained, jump procedure-aware TAS annotation and evaluates TAS performance using 3D poses as input features.

The most common datasets for 3D pose estimation capture daily activities such as walking and talking \cite{h36m, panoptic, 3dpw}. Several datasets \cite{SportsPose, ASPset} capture sports movements, but no published figure skating 3D pose datasets are available for research (for details, see the next section \ref{related-works}). Therefore, we created a 3D pose dataset of complex and dynamic figure skating jumps, FS-Jump3D, with optical markerless motion capture. 
We train the 3D pose estimation model using FS-Jump3D dataset with other primary datasets \cite{h36m, 3dhp, AIST++} to adapt the model to the unique motions of figure skating jumps.

The purpose of this study is to create a 3D pose dataset for understanding figure skating jumps and to explore the usefulness of various input features for TAS, including 3D pose, with our fine-grained annotation dataset as illustrated in Figure \ref{fig:overview}. The contributions of this study are:
(1) Creating the FS-Jump3D dataset of complex and dynamic figure skating jumps using optical markerless motion capture,
(2) Proposing a new fine-grained figure skating jump TAS dataset annotation method with which TAS models can learn jump procedures.
(3) Validating the usefulness of 3D pose features as input and the fine-grained dataset for the TAS model in figure skating.
The FS-Jump3D dataset will be released to add new motion variations to existing 3D pose datasets for research purposes.

\section{Related Work} \label{related-works}
\textbf{Video understanding of human motion in sports.}
In sports, quantitatively evaluating the movements of athletes is crucial. Various tasks such as action prediction \cite{soccer_prediction_2022, soccer_prediction_2024, badmiinton_prediction}, action recognition (classification) \cite{soccer_var, FenceNet, fitness_recognition, basketball_recognition, hockey_recognition_2023, hockey_recognition_2024, badminton_recognition, rugby_recognition}, action detection \cite{rugby_detection}, action localization (spotting) \cite{e2e-spot, soccer_spotting_2023, soccer_spotting_2024, T-DEED}, action assessment \cite{AQA, AQA7, MTL-AQA, AQA_2024}, and TAS \cite{MCFS, SkatingVerse} have been studied to understand human actions from video footage across different sports. Traditional approaches often utilize image features extracted by CNNs and optical flow methods \cite{I3D, SoccerNet, soccer_i3d, baseball_i3d_1, baseball_i3d_2}. However, current research increasingly focuses on techniques such as human detection \cite{yolo, yolov8} and tracking \cite{BoT-SORT, ByteTrack} and 2D and 3D pose estimation to extract temporal information from videos \cite{AutoSoccerPose, fitness_recognition, badminton_recognition, badmiinton_prediction}. The primary reason for this shift is that image features and optical flow are susceptible to background interference and changes in camera perspectives. Understanding human actions in sports involves considering movements at the group level \cite{basketball_recognition, hockey_recognition_2023, hockey_recognition_2024}, joint-level actions at the individual level \cite{suzuki2024automatic, tanaka2023automatic_mmsports}, and the movement of objects like balls in ball games \cite{bascketball_3d_ball_localization, basketball_spin, tabletennis_spin, soccer_2d_ball_localization, badminton_shuttle}. With advancements in tracking and pose estimation technologies, video understanding is expected to be improved in various tasks.

\begin{table*}[ht]
    \centering
    \begin{tabular}{lccccccccc}
        \hline
        Dataset & Markerless & Sync & Subjects & Poses & Keypoints & Environment & Cameras & Frames & FPS \\
        \hline
        Human3.6M \cite{h36m} & $\times$ & hw & 11 & 900K & 26 & lab & 4 & 3.6M & 50 \\
        MPI-INF-3DHP \cite{3dhp} & $\checkmark$ & hw & 8 & 93K & 28 & lab \& outdoor & 14 & 1.3M & 25/50 \\
        3DPW \cite{3dpw} & $\times$ & sw & 7 & 49K & 24 & lab \& outdoor & 1 & 51K & 30 \\
        HumanEva-I \cite{Humaneva} & $\times$ & sw & 6 & 78K & 15 & lab & 7 & 280K & 60 \\
        HumanEva-II \cite{Humaneva} & $\times$ & hw & 6 & 3K & 15 & lab & 4 & 10K & 60 \\
        TotalCapture \cite{TotalCapture} & $\times$ & hw & 5 & 179K & 25 & lab & 8 & 1.9M & 60 \\
        CMU Panoptic \cite{panoptic} & $\checkmark$ & hw & 8 & 1.5M & 18 & lab & 31 & 46.5M & 30 \\
        AIST++ \cite{AIST++} & $\checkmark$ & sw & 30 & 1.1M & 17 & lab & 9 & 10.1M & 60 \\
        ASPset-510 \cite{ASPset} & $\checkmark$ & sw & 17 & 110K & 17 & outdoor & 3 & 330K & 50 \\
        SportsPose \cite{SportsPose} & $\checkmark$ & hw & 24 & 117K & 17 & lab \& outdoor & 7 & 1.5M & 90 \\
        FS-Jump3D (ours) & $\checkmark$ & hw & 4 & 6.5K & 86 & ice rink & 12 & 78K & 60 \\
        \hline
    \end{tabular}
    \caption{Comparison of our FS-Jump3D dataset with other major 3D pose datasets. System configuration (markerless, hardware synchronization, high frame rate) allows capturing high-performance data, including triple jumps, in the unique ice skating rink environment.}
    \label{tab:datasets}
\end{table*}

\textbf{Figure skating video understanding.}
In figure skating, much research has been conducted on tasks such as action recognition \cite{VPD, FSD-10, SkatingVerse}, action localization (detection, spotting) \cite{e2e-spot}, action assessment \cite{AQA, fs_score, AudioVisualMLP} and TAS \cite{MCFS, SkatingVerse}. Numerous approaches have been developed for the Action Assessment task to automate the scoring of entire programs by learning the relationship between past competition footage and judge scores \cite{AQA, fs_score, AudioVisualMLP}. However, these approaches often need more transparent scoring criteria and reliability as evaluation metrics. Consequently, recent studies focus on partially automating the scoring process used by judges. Examples include tasks that recognize the type of jumps from videos \cite{AutoTS, FSD-10, VPD}, automate edge error detection (a technical judgment for jumps) \cite{tanaka2023automatic_gcce, tanaka2023automatic_mmsports}, and determine under-rotations\cite{hirosawa2020}. Additionally, TAS, which assigns action labels to each frame based on the type and rotation count of jumps and spins from untrimmed videos, has been researched to automate the tasks of replay operators and technical specialists \cite{MCFS, SkatingVerse}. Compared to other sports, figure skating does not involve group actions or ball movements, making individual temporal actions a significant clue for video understanding. The FSD-10 \cite{FSD-10} study addresses the classification of the top 10 most frequent jumps and spins from a figure skating competition video dataset, using optical flow as a feature and enhancing performance by extracting keyframes with significant changes in arm and leg keypoints. Research on MCFS \cite{MCFS} and SkatingVerse \cite{SkatingVerse} focuses on the TAS task for figure skating jumps and spins, proposing hierarchical annotation datasets. Experiments with MCFS \cite{MCFS} revealed that using 2D pose sequences as features improved model performance. However, no research has yet explored using 3D pose sequences as input for TAS models, nor has there been any attempt to annotate jumps considering their steps (e.g., ``entry'' and ``landing''). Therefore, this study proposes a detailed annotation dataset considering jump steps and examines the performance of TAS models using 3D pose sequences as input.

\textbf{3d pose estimation dataset.}
Many major 3D pose estimation datasets include everyday movements, such as Human3.6M \cite{h36m}, 3DPW \cite{3dpw}, MPI-INF-3DHP \cite{3dhp}, and HumanEva-I \cite{Humaneva}. Table \ref{tab:datasets} summarizes the characteristics of these major 3D pose estimation datasets. Most traditional benchmark datasets \cite{h36m, 3dhp, Humaneva, TotalCapture} acquire data through motion capture systems requiring markers. While these systems provide highly accurate motion recordings, they cannot capture the dynamic movements of sports because they restrict the subjects' physical movements. In contrast, with the adoption of markerless systems, datasets have been developed that capture dynamic motions such as kicking in soccer, throwing in baseball \cite{SportsPose}, jumping, catching \cite{ASPset}, and dancing \cite{AIST++}. These datasets have enabled 3D pose estimation of body movements that conventional everyday movement-based datasets could not address. However, no dataset captures 3D poses in figure skating jumps within the unique environment of a skating rink. Figure skating movements are challenging to estimate with conventional datasets because the skating rink allows for unique poses utilizing inertia and centrifugal force. Therefore, this study aims to create the FS-Jump3D dataset, which captures figure skating 3D jump poses using optical markerless motion capture.

\section{Proposed Framework}
This study aims to perform Temporal Action Segmentation (TAS) of figure skating jumps using 3D pose sequence features as input. Although previous studies \cite{MCFS, skeleton-based_TAS_1, skeleton-based_TAS_2, skeleton-based_TAS_3, skeleton-based_TAS_4} used 2D poses as input features, the proposed method uses 3D poses, considering that figure skating is a three-dimensional motion. However, figure skating jumps are complex and dynamic motions. The 3D pose estimation model, which has been pre-trained on the primary dataset \cite{h36m, panoptic, cocodataset, 3dpw}, consisting of daily life movements, has limitations in estimation performance for figure skating jump movements \cite{tanaka2023automatic_mmsports}. 
Therefore, in this study, we adapt the 3D pose estimation model for figure skating jump motions by creating the FS-Jump3D dataset, which consists of figure skating jump data obtained through optical markerless motion capture.
In addition, we added detailed annotations on the preliminary jump and landing motion to the TAS dataset. This novel annotation facilitates the model's understanding of figure skating jumps by leveraging the performance of previous TAS models, which can learn ``procedures'' of human actions.
Figure \ref{fig:overview} shows an overview of our proposed method. 
It consists of two steps: training the 3D pose estimation model using the FS-Jump3D dataset along with other primary datasets \cite{h36m, 3dpw, AIST++}, and performing TAS with time-series pose features as input.
First, to create FS-Jump3D dataset, we measured six different types of jumps at a skating rink using optical markerless motion capture. 
Then, we train the 3D pose model using our FS-Jump3D dataset along with three other diverse datasets \cite{h36m, 3dhp, AIST++}.
After that, we estimate the 3D poses from the broadcast video dataset.
Finally, we perform TAS on the dataset annotated with detailed figure skating jump procedures using the extracted 3D poses with normalization and pose alignment processed as input features.

\subsection{FS-Jump3D dataset}
Using optical markerless motion capture (Tehia3D, Tehia), we create a 3D pose dataset of complex and dynamic jump movements from figure skating. We use high-speed cameras (Miqus Video, Qualisys) with hardware synchronization to capture videos for measurement. By installing 12 cameras on the ice skating rink to enable wide-area filming, it is possible to capture not only the jump moment but also the entire jump from entry to landing, which is essential for the TAS task. In addition, the hardware-synchronized markerless method enables the measurement of high-difficulty level jumps, including triple jumps, with less than 1mm error without restricting the skater's movements. In this way, we create FS-Jump3D dataset, which consists of 12 video viewpoints and 3D jump poses. This dataset contains the position coordinates of 86 keypoints (head: 6, torso: 16, arms: 30, legs: 34) and can also be used for more detailed motion analysis. FS-Jump3D will play a role in developing 3D pose estimation technology in sports, where data diversity is required.

\subsection{Optical Markerless Motion Capture}
\subsubsection{Setup}
\begin{figure}[ht]
    \centering
    \includegraphics[width=0.9\linewidth]{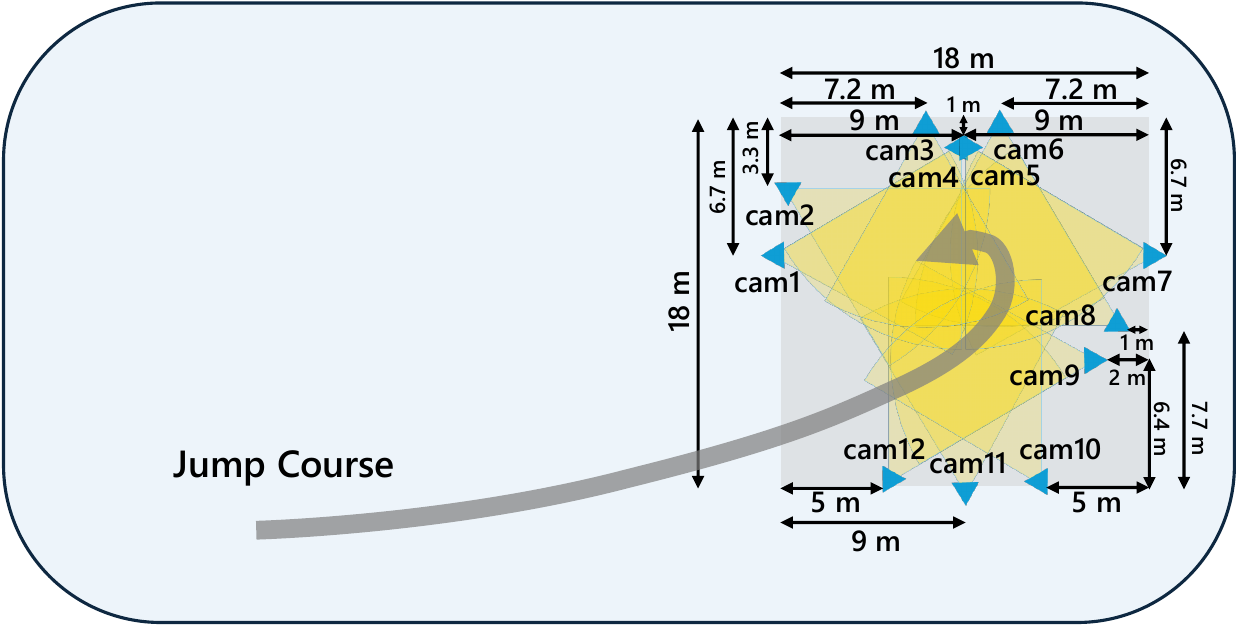}
    \caption{Layout of 12 high-speed cameras on the ice skating rink.}
    \label{fig:camera_position}
\end{figure}

We placed 12 high-speed cameras to capture half of the ice skating rink, as shown in Figure \ref{fig:camera_position}. The camera layout considered the jump course and did not restrict the skaters' jump movements as much as possible. We calibrated the cameras with dedicated software (Qualisys Track Manager, Qualisys) and a wand-type calibration kit. The cameras had 2M pixels, and for each jump shot, the frame rate was 60 fps, the shooting time was 5 seconds (300 frames), and the resolution was $1920 \times 1080$ pixels.

\subsubsection{Participants}
Four skaters with sufficient figure skating experience participated in capturing our FS-Jump3D dataset (after this, they will be called Skater A-D). In the badge test, a proficiency test set by the Japan Skating Federation, Skater A was at level 7, Skater B was at level 5, and Skaters C and D were at level 6. For the five types of jumps except for the Axel, Skater A attempted a triple jump, and Skaters B to D attempted a double jump. For the Axel jump, Skaters A and C attempted double Axels, and Skaters B and D attempted one and single Axels.

\subsubsection{Capturing Conditions}
We captured ten attempts of all six types of jumps (Axel, Salchow, Toe Loop, Loop, Flip, and Lutz) for each skater. Through ten attempts of each type of jump, we counted each attempt as one, even if the skater missed or fell over the jump. Missed jumps and falls often occur in competitions, and the pose information is essential for judging, such as under-rotations. Therefore, the data on jump misses and falls captured in FS-Jump3D are practical as more natural data that is close to the actual situation. In addition to the six types of jump, three or four attempts of combination jumps, such as Axel $+$ Toe Loop, were filmed for each skater. By processing these 12 views of video data using Theia3D, we finally created FS-Jump3D dataset consisting of a total of 253 videos and 3D pose data.

\subsection{Training 3D Pose Estimation Model with FS-Jump3D}\label{training}
We train a 3D pose estimation model using the FS-Jump3D dataset. For the 3D pose estimation model, we use a Transformer-based approach and a graph-based method called MotionAGFormer \cite{motionagformer2024}. Besides the FS-Jump3D dataset, to address the variety of figure skating motions in broadcast videos, we train the model with three additional datasets: Human3.6M \cite{h36m} (daily motions), MPI-INF-3DHP \cite{3dhp} (daily motions and outdoor activities), and AIST++ \cite{AIST++} (dance). We examined various usage of dataset, which is described in Section \ref{sec:experiment}. For all datasets, we employed the Human3.6M keypoint rig. Additionally, to enable the model to focus on learning non-rigid human joint movements, we centered the pose at the root joint (hip midpoint) and normalized the joint position coordinates by the maximum absolute value for each sequence. Finally, we trained the model with these four datasets, adapting it to figure skating jump motions.

\subsection{Fine-Grained and Jump Procedure-Aware Annotation}
We perform TAS of figure skating jumps using time-series pose features. Our dataset consists of broadcast videos from the Olympics and World Championships, previously used in studies on figure skating jump classification and action spotting \cite{VPD, e2e-spot}. This figure skating video dataset consists of broadcast footage from a total of 371 men's and women's short programs.
In conventional figure skating TAS datasets \cite{MCFS, SkatingVerse}, segments encompassing dozens of frames before and after the takeoff and landing of a jump are annotated with a single action label. However, these annotations do not precisely indicate the timing of takeoff and landing, nor do they allow the model to learn the jump procedure. In figure skating, the preparation motion is crucial for identifying the type of jump, as each jump requires distinct body movements to generate rotation. In addition, the landing motion is a critical cue for pinpointing the moment of ice landing. Therefore, we propose a novel fine-grained annotation for each jump, considering figure skating jumps as three phases: ``entry'', ``jump (takeoff to land)'', and ``landing''. This annotated dataset allows the model to understand the jump procedure and segment it precisely by clearly defining the moment of takeoff and landing. We label jump ``entry'' starting three steps before takeoff, considering turns or skating leg swaps as step changes. We define the ``landing'' as the duration the back outside edge skating sustains. While the entry phase is unique for each six types of jump (e.g., ``Axel entry'' and ``Salchow entry''), the landing phase is common to all jumps. In addition, as in previous studies \cite{SkatingVerse}, we introduced a set level and element level hierarchy in the annotation of the ``jump'' label; in the set level annotation, each jump is classified into six labels by its type (e.g., ``Salchow jump'', ``Axel jump''). In the element level annotations, each jump is classified into 23 labels by its type and rotation (e.g. ``4 Salchow jump'', ``3 Axel jump''). Consequently, element level TAS is more challenging than set level TAS, as it requires not only the identification of jump types but also the rotation. Including annotations for ``entry'' and ``landing'', the final dataset comprises 13 action labels at the set level and 30 action labels at the element level. (Note that all other frames not involving jump actions are assigned the label ``NONE''.)

\subsection{Figure Skating Jump TAS}
\begin{figure}[ht]
    \centering
    \includegraphics[width=0.95\linewidth]{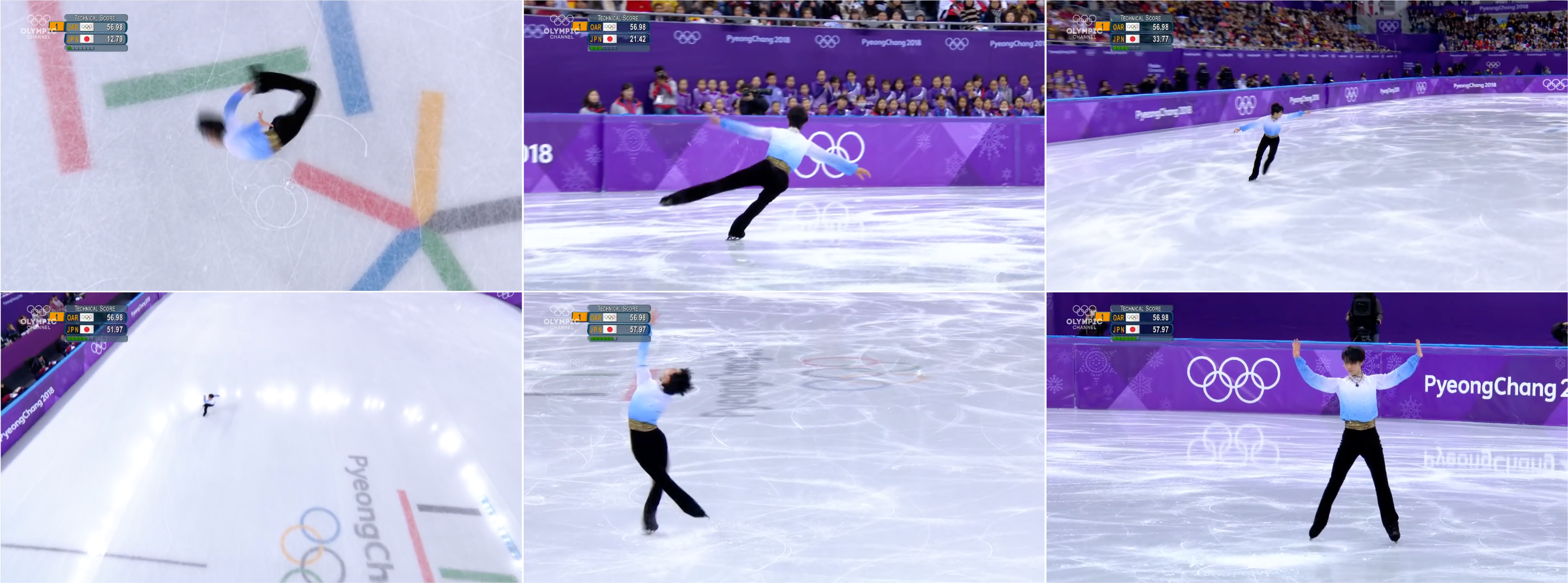}
    \caption{Examples of the variety of shooting angles in figure skating broadcast footage.}
    \label{fig:shooting_ang}
\end{figure}

For the TAS model, we use FACT \cite{FACT}, a Transformer-based method. This model employs a framework in which the frame- and action-level characteristics are mutually learned through cross-attention and has achieved the best performance in four primary TAS datasets \cite{Breakfast, GTEA, Ego-ProceL, Epic-Kitchen}. We use 2D and 3D poses estimated from a broadcast video dataset as input features. First, we estimate the 2D pose of 17 joints and the confidence scores for each joint by DWPose \cite{DWPose}, which has been pre-trained on the COCO Wholebody dataset \cite{coco-wholebody}. 
Next, using the extracted time-series 2D pose data as input, we estimate the 3D pose by MotionAGFormer \cite{motionagformer2024} trained on our FS-Jump3D dataset along with other primary datasets \cite{h36m, 3dhp, AIST++} (see section \ref{training}). 
At this time, the 2D inputs and 3D outputs of MotionAGFormer were zero-masked if the keypoint's confidence score of DWPose was below 0.3. For further processing of the 3D poses, we applied a rotation to align the estimated 3D pose to face the same direction throughout the sequence (pose alignment) and included the Euler angles comprising the applied rotation matrix as additional inputs for each frame. As shown in Figure \ref{fig:shooting_ang}, figure skating broadcast footage is shot from various angles. Therefore, by applying these pose-alignment processes, we create features that are independent of the shooting angle. In addition to these 2D and 3D poses, we also examine 2D-VPD features, which are pose features derived from RGB frames by Video Pose Distillation \cite{VPD}. We verify TAS performance using these time-series 2D poses, 3D poses, 2D-VPD features, and their combinations as input to FACT.

\section{Experiments}\label{sec:experiment}
\subsection{Datasets}
\subsubsection{FS-Jump3D Dataset}
To compare the characteristics of our FS-Jump3D dataset to other primary datasets, we provide an overview in Table 1. The most important feature of our dataset was its shooting environment. While conventional datasets are taken either in a lab (indoors) or outdoors, the FS-Jump3D dataset was taken at a skating rink. Unlike other environments, the skating rink has very little friction with the ground for a person wearing bladed skates, which allows for complex poses using centrifugal force. All six jump poses in the FS-Jump3D dataset cannot be reproduced on land, making our on-ice filming valuable from a rarity perspective. Our FS-Jump3D dataset is the first publicly available dataset that captures the 3D poses of figure skating jumps. In addition, we created FS-Jump3D dataset with a high frame rate, markerless, and hardware synchronization. A markerless system does not restrict the athlete's movement, and hardware synchronization can accommodate excessive joint movement between frame exposures. These system settings make it possible to capture high-performance data, including five different types of triple jumps, which is a strength of this dataset. Although the number of frames of pose data was inferior to other datasets, the number of cameras was relatively large, providing viewpoints from various angles. Although not implemented in this study, this video could be utilized to create a 2D pose dataset for the complex jumping poses of figure skating, where occlusion frequently occurs.

\subsubsection{Figure Skating TAS Dataset} \label{TAS_anno_stats}
We present some statistics on our annotated TAS dataset. The average number of total frames per video is 4265, of which 382 frames are assigned action labels (from jump entry to landing). This results in a ratio of action labels to total frames of 8.96\%, indicating the challenge of the TAS task for figure skating jumps with competition videos as input.

\begin{figure}[ht]
    \centering
    \includegraphics[width=1.0\linewidth]{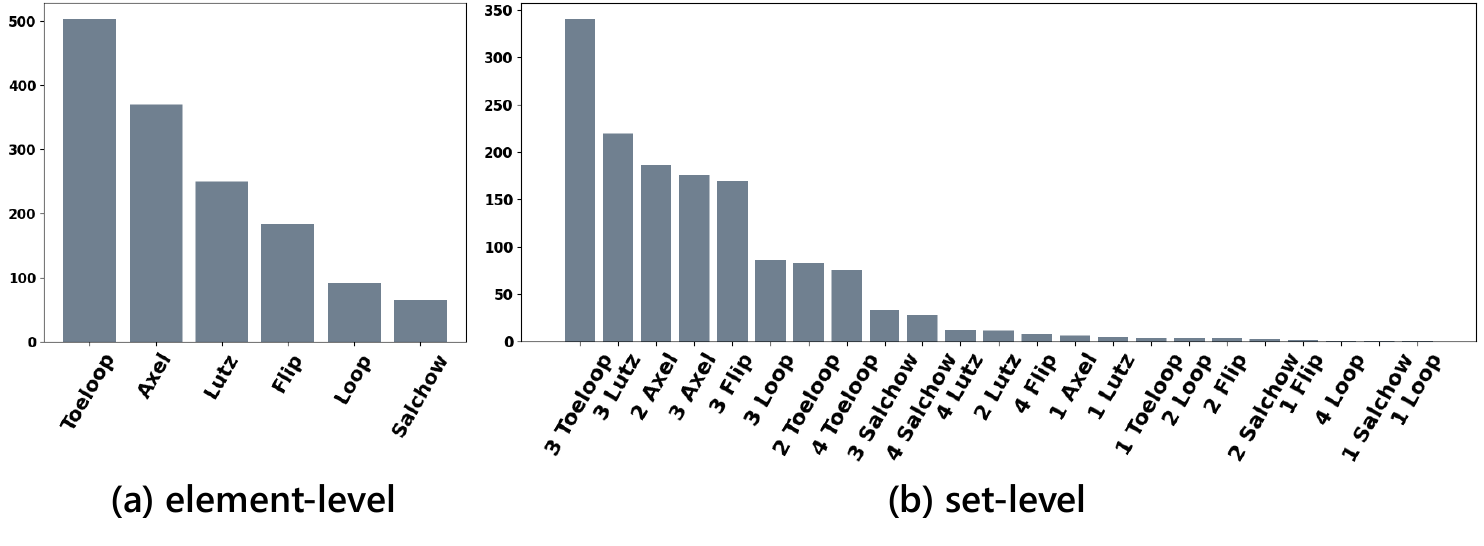}
    \caption{The number of occurrences of each jump in all 371 performance videos we annotated.}
    \label{fig:tas_anno_stats}
\end{figure}

Figure \ref{fig:tas_anno_stats} summarizes the number of occurrences of each jump in our annotated TAS dataset. Set level annotations (a) show that the Toeloop had the highest number of occurrences because many skaters select Toeloops for solo jumps and as the second jump in combination jumps. The Axel jump had the second-highest number of occurrences. At the element level (b), ``3 Toeloop'' occurs most frequently, followed by ``3 Lutz'', ``2 Axel'', and ``3 Axel''. In the championship class, the short program requires an Axel-type jump, so almost all ladies perform ``2 Axel'', and most men perform ``3 Axel''. Less common are quadruple jumps of high difficulty and single and double jumps resulting from failures. These findings indicate that element level task is more difficult than the set level due to the uneven number of occurrences of each jump, especially at the element level.

To evaluate our proposed fine-grained annotation method, we define a coarse annotation at the set level. In the coarse annotation, the ``entry'' and ``landing'' labeled at the proposal set level annotation are all replaced with ``NONE'' labels, and the only ``jump'' (e.g., ``Salchow jump'', ``Axel jump'') is annotated as the action label. This coarse annotation simulates the MCFS \cite{MCFS} dataset annotation, which does not account for the jump procedure. To maintain consistency in the segmentation evaluation conditions, we create a coarse annotation by omitting the jump procedure annotation from the proposed annotation.

\subsection{Evaluation}
We evaluated the estimation error on the test data of the FS-Jump3D dataset both with and without using the FS-Jump3D dataset for training the 3D pose estimation model. The 3D pose estimation model was trained on three primary datasets (Human3.6M \cite{h36m}, MPI-INF-3DHP \cite{3dhp}, AIST++ \cite{AIST++}) along with the FS-Jump3D dataset. The FS-Jump3D training data consisted of 190 jump sequences from Skaters A-C, while the test data comprised 63 jump sequences from Skater D. For comparison, the 3D pose estimation model pre-trained on the same three datasets \cite{h36m, 3dhp, AIST++} was also validated when fine-tuned on the FS-Jump3D dataset.

As an evaluation metric, we used the Mean Per Joint Position Error (MPJPE) to calculate the average distance between the estimated position of all joints and the ground-truth position.
Next, we evaluated the TAS performance for each input feature for the fine-grained dataset annotated with our method. The way of splitting the dataset was the same as in the previous study \cite{e2e-spot}, with all 2018 competition videos used for testing and the rest for train and validation. The 2D-VPD \cite{VPD} used as input features is an image feature and is therefore affected by the background of the competition venue and the camera's shooting angle. Therefore, we utilize a competition-by-competition split method to evaluate the generalization performance for unknown competition venues. We compute accuracy and F1@\{10, 25, 50, 75, 90\} for all action segments to evaluate the TAS task as in previous studies \cite{TAS_analysis}. Accuracy measures the proportion of correctly classified frames over the total number of frames. F1@\{10, 25, 50, 75, 90\} computes the harmonic mean of precision and recall, where we consider predicted segments correct if they overlap with the ground truth by at least \{10, 25, 50, 75, 90\}\%.

\subsection{Results}
\subsubsection{3D pose estimation performance with FS-Jump3D Dataset}
Table \ref{table:mpjpe} compares MPJPE with and without using the FS-Jump3D dataset for training 3D pose estimation models. The results show that the MPJPE is significantly reduced when the FS-Jump3D dataset is included in the training process. 
Our primary objective is to enhance TAS performance on the figure skating broadcast video dataset, which includes not only jumps but also other movements.  While MPJPE is lowest when the FS-Jump3D dataset is used for fine-tuning, to prevent overfitting, we employ a method that incorporates the FS-Jump3D dataset with other datasets during training from scratch.

\begin{figure}[h]
    \centering
    \includegraphics[width=0.9\linewidth]{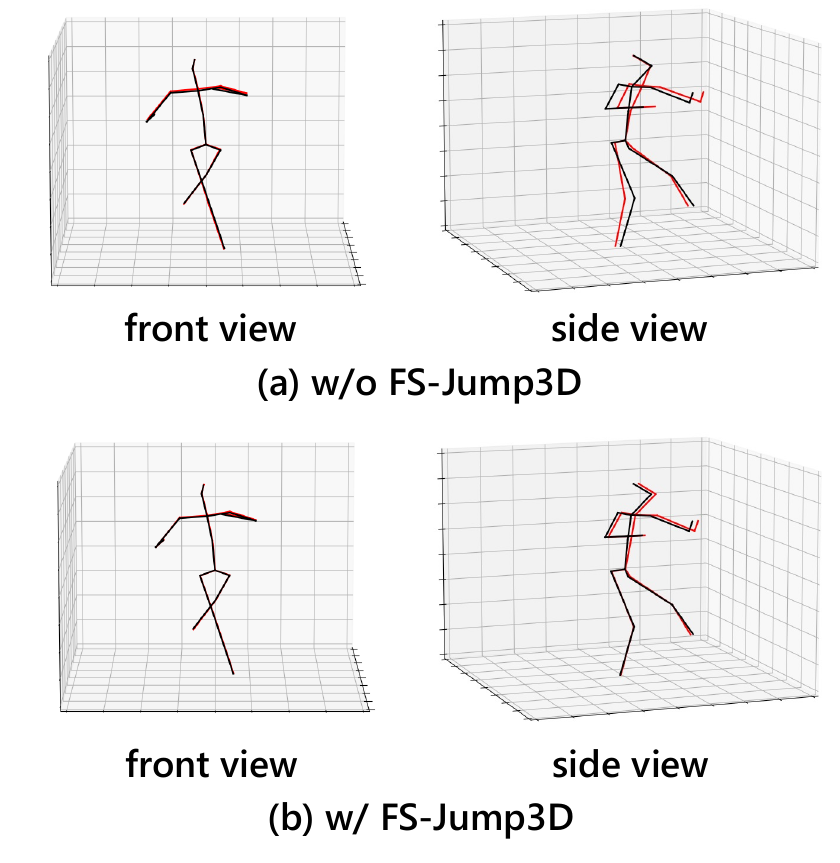}
    \caption{Visualization of estimation results with and without the FS-Jump3D dataset for training a 3D pose estimation model.}
    \label{fig:vis}
\end{figure}

Figure \ref{fig:vis} displays the estimated results for the FS-Jump3D test data from the same viewpoints and sides as the input 2D images, both with and without using the FS-Jump3D dataset for training the model. The figure illustrates the moment of landing during a double Lutz jump. It demonstrates that the estimation results from the same orientation as the 2D input have low error whether the FS-Jump3D dataset is used or not. However, in the side view (depth direction of the 2D input), there is a noticeable improvement when the FS-Jump3D dataset is used for training. These results indicate that incorporating the FS-Jump3D dataset into the training data enhances the 3D pose estimation for figure skating jumps.

\begin{table}[ht]
\centering
\begin{tabular}{lcc}
\hline
\textbf{Trainig method} & \textbf{MPJPE} \\ \hline
Scratch (w/o FS-Jump3D) & 55.432 mm \\
Scratch (w/ FS-Jump3D)& 24.963 mm \\
Fine-tuning (w/ FS-Jump3D) & 17.357 mm \\ \hline
\end{tabular}
\caption{MPJPE of the 3D pose estimation model with and without FS-Jump3D dataset added to the other three dataset \cite{h36m, 3dhp, AIST++}. The test set consists of data from Skater D of the FS-Jump3D dataset.}
\label{table:mpjpe}
\end{table}

\subsubsection{Figure Skating Jump TAS with our Fine-Grained Dataset}

\begin{table*}[ht]
\centering
\begin{tabular}{lcccccc}
\hline
Feature & Acc & F1@10 & F1@25 & F1@50 & F1@75 & F1@90 \\
\hline
2D pose (Baseline) & 78.55 & 85.12 & 84.93 & 84.17 & 81.52 & \textbf{35.83} \\
3D pose (w/o FS-Jump3D) & 75.76 & 83.08 & 82.89 & 82.51 & 77.57 & 31.18 \\
3D pose & \textbf{79.89} & \textbf{87.13} & \textbf{86.94} & \textbf{86.56} & \textbf{82.36} & 33.36 \\
3D pose (w/o p.a. \& e.a.) & 78.79 & 85.00 & 84.81 & 84.05 & 81.18 & 34.96 \\
\hline
2D-VPD (Baseline) & 81.25 & 86.42 & 86.42 & 86.23 & 85.09 & \textbf{44.82} \\
2D-VPD+3D pose & 81.23 & 88.02 & 88.02 & 87.64 & 85.36 & 40.49 \\
2D-VPD+3D pose (w/o p.a. \& e.a.) & \textbf{83.10} & \textbf{89.14} & \textbf{89.14} & \textbf{88.95} & \textbf{86.67} & 41.52 \\
\hline
\end{tabular}
\caption{Comparison of TAS performances with various input features in the set level.}
\label{tab:set_level}
\end{table*}

\begin{table*}[ht]
\centering
\begin{tabular}{lcccccc}
\hline
Feature & Acc & F1@10 & F1@25 & F1@50 & F1@75 & F1@90 \\
\hline
2D pose (Baseline) & \textbf{71.34} & \textbf{78.97} & \textbf{78.97} & \textbf{78.78} & \textbf{75.74} & \textbf{35.39} \\
3D pose (w/o FS-Jump3D) & 70.49 & 78.36 & 78.17 & 77.98 & 72.07 & 29.17 \\
3D pose & 70.17 & 77.71 & 77.33 & 76.57 & 71.62 & 29.52 \\
3D pose (w/o p.a. \& e.a.) & 68.22 & 74.60 & 74.60 & 74.21 & 71.55 & 32.35 \\
\hline
2D-VPD (Baseline) & \textbf{78.85} & \textbf{84.46} & \textbf{84.46} & \textbf{84.27} & \textbf{82.94} & \textbf{47.85} \\
2D-VPD+3D pose & 76.06 & 81.75 & 81.56 & 81.56 & 80.23 & 39.92 \\
2D-VPD+3D pose (w/o p.a. \& e.a.) & 72.63 & 77.91 & 77.72 & 77.35 & 76.21 & 39.62 \\
\hline
\end{tabular}
\caption{Comparison of TAS performances with various input features in element level.}
\label{tab:element_level}
\end{table*}

Tables \ref{tab:set_level} and \ref{tab:element_level} compare TAS results for figure skating jumps using various input features at the set and element levels. The upper part of each table shows the case where the joint position coordinates obtained from pose estimation are used as input features, with the 2D pose set as the baseline. The lower part of each table shows the case where 2D-VPD obtained from a previous study is utilized, with 2D-VPD set as the baseline. In these tables, "w/o p.a.” indicates that the 3D pose is processed without pose-alignment, and "w/o e.a." denotes 3D pose features that do not include the Euler angles used in pose-alignment. The comparison between Tables \ref{tab:set_level} and \ref{tab:element_level} reveals that TAS at the element level is more challenging than at the set level.

We examined the TAS performance of each feature at the set level in Table \ref{tab:set_level}. First, we consider the case where joint position coordinates in the upper half of Table \ref{tab:set_level} are used as input: a comparison of 2D pose and 3D pose indicates that the 3D pose was the more important feature in the TAS of figure skating jumps when joint position coordinates were used as input at the set level. In addition, the comparison between 3D pose (w/o FS-Jump3D) and 3D pose confirms that the TAS performance of figure skating jumps can be improved by utilizing the FS-Jump3D dataset created in this study for training 3D pose estimation models. This indicates that it would be important to capture competition-specific motions more accurately in TAS that takes joint position coordinates as input to the model. The ablation test of the pose-alignment process and the addition of the Euler angle to the features used in the process suggest that the addition of pose-alignment and Euler angles would be useful at the set level. This suggests that while the 2D pose is a feature that depends on the shooting angle of the broadcast video, the pose-aligned 3D pose is a feature that does not depend on the shooting angle, and thus can be a better input for the TAS.

Next, we investigated the case where the 2D-VPD features in the lower half of Table \ref{tab:set_level} are utilized. At the set level, TAS performance was improved when 2D-VPD was used as an input to the model in combination with the 3D pose rather than as an input alone. However, it is suggested that the 3D pose without pose-alignment processing may be slightly more useful when used in combination with 2D-VPD. Since 2D-VPD is an embedded representation, it is robust to the shooting angle of broadcast video, and a pure 3D pose without pose-alignment processing may be useful for use in combination.

Next, we examined the TAS performance of each feature at the element level, as shown in Table \ref{tab:element_level}. Among the cases where we use joint position coordinates as input features, the 2D pose input demonstrates the best performance at the element level. Additionally, when utilizing 2D-VPD, the highest performance was achieved when 2D-VPD was used alone rather than in combination with the 3D pose. These results indicate challenges in using 3D pose for TAS of figure skating considering jump rotation. Due to the high speed of figure skating rotations, some 3D pose estimation results show that the rotations were smoothed out in the frame sequence when lifting the 2D pose estimation to 3D. This indicates that the 3D pose estimation model developed in this study has issues accurately estimating figure skating jump rotations. When training the 3D pose estimation model, we used the FS-Jump3D dataset by downsampling it to 30 fps while the broadcast video was at 25 fps. However, there was a difference between the rotation speeds of double and triple jumps in the training data, and those of triple and quad jumps in the broadcast video. This discrepancy in rotation speeds may have caused some jumps to be inaccurately estimated. It is also possible that the number of datasets used in this study would be insufficient for the TAS of the element level; as mentioned in section \ref{TAS_anno_stats}, the datasets used in this study are highly biased for each jump element. Future work could include the FS-Jump3D dataset considering differences in jump rotation speed and fps between the 3D pose estimation datasets and TAS datasets, and expanding the TAS datasets.

In both the set and element level evaluation, the evaluation result by F1@90 is significantly lower than that by F1@10 to F1@75. Given that the average jump duration (from takeoff to land) is 16.25 frames, F1@90 requires approximately 15 frames of overlap between ground truth and TAS results. This requirement indicates that F1@90 is a rigorous evaluation metric, allowing for an error margin of only about 1 or 2 frames. In contrast, F1@75 requires an overlap of about 12 frames, allowing for an error margin of approximately 4 frames. This index shows results exceeding 80\% for both set-level and element-level validation. In this study, we proposed a fine-grained annotation based on jump takeoff and land. We found that while F1@90 represents the upper limit of the model's performance, the model can reliably achieve high performance under the F1@75 metric.

In Table \ref{tab:annotation_comparison}, we present the results of testing the effectiveness of our proposed annotations for improving the TAS model's understanding of figure skating jump procedures. We compared the set level annotation with the coarse annotation in which ``entry'' and ``landing'' are labeled as ``NONE'' and only ``jump'' as a valid action label. For the evaluation metric, we used F1@50, which is the most common in TAS tasks. Although direct comparison is impossible, the results indicate that the proposed annotations yield better TAS performance considering the jump procedure. For the coarse annotation, action labels are assigned to 1.50\% of the total frames, whereas action labels are 8.96\% of the total frames in the proposed annotation. This difference between action label ratios suggests that the proposed annotation method can potentially enhance the TAS performance for figure skating jumps by assigning more meanings to frames and considering the jump procedure.

\begin{table}[ht]
\centering
\begin{tabular}{lccc}
\hline
Feature & Proposal & Coarse \\
\hline
2D pose & \textbf{84.17} & 76.42 \\
3D pose (w/o FS-Jump3D) & \textbf{82.51} & 73.76 \\
3D pose & \textbf{86.56} & 72.78 \\
3D pose (w/o p.a. \& e.a.) & \textbf{84.05} & 73.13 \\
\hline
2D-VPD & 86.23 & \textbf{86.35} \\
2D-VPD+3D pose & \textbf{87.64} & 82.08 \\
2D-VPD+3D pose (w/o p.a. \& e.a.) & \textbf{88.95} & 82.62 \\
\hline
\end{tabular}
\caption{Comparison of F1@50 at a set level between the proposed and coarse annotations.}
\label{tab:annotation_comparison}
\end{table}


Table \ref{tab:scratch_vs_fine-tuning} presents a comparison of the proposed method using the FS-Jump3D dataset from scratch and fine-tuning. In our proposed method, the FS-Jump3D dataset and three other datasets \cite{h36m, 3dhp, AIST++} were simultaneously trained for the 3D pose estimation model to accommodate the variety of movements (i.e., to avoid overfitting to FS-Jump3D dataset) in figure skating broadcast videos. In this scenario, the MPJPE for the test data of the FS-Jump3D dataset was 24.96 mm. Conversely, when the FS-Jump3D dataset was used for fine-tuning a pre-trained model with the three other datasets \cite{h36m, 3dhp, AIST++}, the MPJPE for the FS-Jump3D test data was 17.35 mm (see Table \ref{table:mpjpe}). To determine whether the FS-Jump3D dataset should be used for training from scratch or for fine-tuning, we evaluated TAS performance using the 3D poses obtained from each method as input to the TAS model.
The results showed little difference in TAS performance at both the set level and element level, regardless of the training method used (see Table \ref{tab:scratch_vs_fine-tuning}). In this study, we used FS-Jump3D dataset for training from scratch with other datasets to prevent overfitting; however,  further investigation is required to determine the most effective way to utilize the FS-Jump3D dataset.

\begin{table}[htbp]
\begin{subtable}[t]{0.45\textwidth}
\centering
    \begin{tabular}{lccc}
    \hline
    Feature & Acc & F1@50\\
    \hline
    3D pose (scratch) & 79.89 & \textbf{86.56} \\
    3D pose (fine-tuning) & \textbf{80.27} & 86.33 & \\
    \hline
    \end{tabular}
    \caption{set level}
    \label{tab:set}
\end{subtable}%
\hfill
\begin{subtable}[t]{0.45\textwidth}
\centering
    \begin{tabular}{lccc}
    \hline
    Feature & Acc & F1@50\\
    \hline
    3D pose (scratch) & \textbf{70.17} & 76.57 \\
    3D pose (fine-tuning) & 69.18 & \textbf{76.76} & \\
    \hline
    \end{tabular}
    \caption{element level}
    \label{tab:element}
\end{subtable}%
\caption{Comparison of TAS performance with 3D pose as input features when the FS-Jump3D dataset is used for training from scratch and for fine-tuning.}
\label{tab:scratch_vs_fine-tuning}
\end{table}

\section{Conclusion}
In this study, we validated the TAS of figure skating jumps using 3D poses as input. We also proposed a fine-grained TAS annotation that considers the jump procedure and created FS-Jump3D, a 3D jump pose dataset for figure skating, including triple-rotational jumps. The validation results confirmed that the TAS performance using 3D poses, obtained by training a 3D pose estimation model with the FS-Jump3D dataset, exceeded that of 2D pose input in the set-level TAS task for segmenting jump types.
Additionally, comparing TAS results with coarse annotation, which does not consider the jump procedure, showed that the proposed annotation method is more practical for the TAS task of figure skating jumps. However, the proposed method utilizing the FS-Jump3D dataset did not accurately estimate some of the jump rotations in the broadcast video. This led to inferior performance of the 3D pose as an input feature in the element-level TAS task that considers jump rotation. Future work will involve using the FS-Jump3D dataset, considering the speed of jump rotations and fps in the broadcast video, and expanding the TAS dataset.

\section{Acknowledgment}
We would like to express our gratitude to Ms. Taniguchi and Prof. Sakurai of Chukyo University for their generous support in providing access to the skating rink, which greatly facilitated the collection of high-quality motion capture data for figure skating jumps. We also acknowledge the financial support provided by JSPS KAKENHI under Grant Numbers 23H03282 and 21H05300.

\bibliographystyle{ACM-Reference-Format}
\bibliography{ref}

\end{document}